\documentclass{article}
\usepackage{amsmath}
\usepackage{graphicx}
\usepackage{hyperref}
\usepackage{enumitem} % for better bullet points

\title{Integrating Randomness in Large Language Models: A Linear Congruential Generator Approach for Generating Clinically Relevant Content}
\author{Andrew Bouras, B.S. \\
Nova Southeastern University, College of Osteopathic Medicine}
\date{July 2024}

\begin{document}

\maketitle

\begin{abstract}
The generation of diverse, high-quality outputs from language models (LLMs) is essential for various applications, including education and content creation. Achieving true randomness and avoiding repetition remains a significant challenge. This study employs the Linear Congruential Generator (LCG) method for systematic fact selection, combined with AI-powered content generation. Using LCG, we ensured unique combinations of gastrointestinal physiology and pathology facts across multiple rounds, integrating these facts into prompts for GPT-4o to create clinically relevant, vignette-style outputs. Over 14 rounds, 98 unique outputs were generated, demonstrating LCG's effectiveness in producing diverse and high-quality content. This method addresses key issues of randomness and repetition, enhancing the quality and efficiency of LLM-generated content for various applications.
\end{abstract}

\section{Introduction}

\subsection{Background}

\textbf{Randomness in Large Language Models}\\
Randomness is essential in LLMs for generating diverse content. LLMs rely on stochastic processes to produce varied and human-like text, ensuring outputs are not deterministic. This randomness allows LLMs to generate different responses to the same prompt, enhancing their utility in content creation, conversational agents, and educational tools \cite{brown2020language}. Randomness also helps maintain diversity and quality in text generation \cite{zhang2020trading} and facilitates coherent and contextually appropriate responses for applications like paraphrase generation and text style transfer \cite{du2022diverse}. Techniques like the LCG systematically incorporate randomness, ensuring variability and control in outputs \cite{de2019emergence}.

In generating clinically relevant content, randomness ensures diverse and comprehensive material. Stochastic elements allow LLMs to create varied scenarios and questions, reducing redundancy and enhancing educational value \cite{jin2023matching,malek2023toward}. Algorithms like LCG generate pseudo-random numbers to guide content selection, balancing randomness and control \cite{knuth1997art}. This integration of controlled randomness creates diverse and clinically relevant content, advancing the application of LLMs beyond traditional deterministic approaches.\\

\textbf{Multiple-Choice Questions}\\
Multiple-Choice Questions (MCQs) are crucial in medical education, serving as key tools for both formative and summative assessments. They efficiently evaluate a wide range of knowledge and clinical reasoning skills across medical disciplines \cite{case2002}. MCQs offer objective scoring, comprehensive content coverage, and the ability to assess large cohorts simultaneously \cite{tarrant2012}. Well-constructed MCQs can distinguish between varying levels of student performance, providing insights into learners' comprehension of complex medical concepts \cite{schuwirth2011}.

Creating high-quality MCQs is challenging. Developing questions that are clinically relevant and cognitively demanding requires significant time, expertise, and resources \cite{collins2006}. Educators must craft questions that assess higher-order thinking skills, including application, analysis, and evaluation of medical knowledge \cite{haladyna2013}. Ensuring diversity in question content and preventing repetition across assessments is particularly challenging given the extensive and evolving nature of medical curricula \cite{downing2005}. The rapid pace of medical advancements necessitates continual updates to assessment materials \cite{cook2013}.

\subsection{Objective}

This study investigates integrating controlled randomness into LLMs to generate clinically relevant content, focusing on high-quality MCQs in medical education. By employing the LCG method, we aim to combine algorithmic precision with the adaptive capabilities of AI-powered language models.

Our research addresses the challenge of introducing structured variability into AI-generated content while maintaining clinical accuracy and relevance. Using the LCG method for systematic fact selection and integrating it with advanced language models, we aim to create a framework that produces diverse, cognitively challenging, and clinically pertinent educational materials.

The broader implications extend beyond question generation, exploring how controlled randomness can enhance the versatility and applicability of LLMs in specialized domains like medical education. This approach aims to improve the quality and diversity of assessment materials and provide new methodologies for guiding AI systems in generating content that adheres to specific criteria while maintaining creativity and relevance.

Ultimately, this research seeks to advance AI applications in education by offering a solution that balances algorithmic control with AI adaptability. This approach could transform the development and utilization of educational materials in medical education and beyond.

\section{Methodology}

\subsection{Linear Congruential Generator}

LCG is a fundamental and widely-used pseudo-random number generator (PRNG) algorithm that operates on the principle of linear recurrence. The LCG generates a sequence of numbers using the recurrence relation:
\[
X_{n+1} = (aX_n + c) \mod m
\]

Where:
\begin{itemize}[label=--, itemsep=1em]
    \item \(X_n\) is the current number in the sequence.
    \item \(a\) is the multiplier.
    \item \(c\) is the increment.
    \item \(m\) is the modulus.
\end{itemize}

The choice of parameters \(a\), \(c\), \(m\), and the seed value \(X_0\) is critical for the performance of the LCG. These parameters determine the period, randomness, and statistical properties of the generated sequence. In our study, the following parameters were used:

\begin{itemize}[label=--, itemsep=1em]
    \item \textbf{Seed} (\(X_0\)): 12345
    \item \textbf{Multiplier} (\(a\)): 1103515245
    \item \textbf{Increment} (\(c\)): 12345
    \item \textbf{Modulus} (\(m\)): \(2^{31}\)
\end{itemize}

These parameters were selected based on established practices to balance computational simplicity with the quality of randomness \cite{knuth1997art}. The seed value influences the initial state but does not affect the statistical properties beyond the initial value. The multiplier 1103515245 ensures a long period and good randomness properties, while the increment 12345 avoids trivial sequences. The modulus $2^{31}$ provides a wide range for the output values.

The LCG algorithm with these parameters ensures desirable statistical properties, such as uniform distribution and long period, which are essential for avoiding repetitions and maintaining randomness in simulations and randomized content generation \cite{de2019emergence}.

In this study, the LCG generated sequences of indices to select facts from a predefined pool of 100 clinically relevant facts about gastrointestinal physiology and pathology. Mapping the generated pseudo-random numbers to these indices ensured each round of MCQ generation was based on a unique and diverse set of facts, preventing repetition and promoting comprehensive topic coverage, enhancing the educational value of the generated MCQs \cite{maxrizal2022}.

The LCG's simplicity and ability to produce high-quality random sequences underscore its suitability for educational content generation and other applications requiring efficient and effective random number generation \cite{al2019}.

\subsection{Fact Selection Process}

We created a comprehensive fact mapping containing 100 clinically relevant facts about gastrointestinal physiology and pathology. Each fact was assigned a unique index from 1 to 100. The LCG algorithm was used to generate a sequence of pseudo-random numbers, which were then mapped to these fact indices. The fact selection process adhered to two primary criteria: uniqueness and clinical relevance. A set of used facts was maintained throughout the experiment to ensure that each fact had not been previously used, guaranteeing uniqueness across rounds. This approach is crucial for maintaining the diversity and comprehensiveness of the content generated \cite{knuth1997art}.

Ensuring clinical relevance was paramount; all facts in the mapping were curated to ensure their relevance to gastrointestinal medicine, covering a wide range of topics from basic physiology to complex pathological conditions. This curated approach helps in generating content that is not only diverse but also highly pertinent to the medical field \cite{jin2023matching}. The process of fact selection using LCG was methodically structured to avoid biases and ensure a balanced representation of different clinical scenarios, thereby enriching the educational and practical value of the generated content \cite{malek2023toward}.

By systematically managing and directing the randomness through the LCG algorithm, we were able to achieve a high degree of control over the selection process, thus maintaining the integrity and quality of the educational content produced \cite{manathunga2023}. This innovative approach underscores the potential of combining algorithmic precision with large language models to enhance the generation of clinically relevant content.

\subsection{MCQ Generation}

For generating MCQs, we utilized a carefully crafted base prompt. This prompt instructed the AI model to create clinically relevant, vignette-style questions that required the application of knowledge rather than simple recall. The prompt specified the creation of seven high-quality MCQs per round, including patient demographics, presenting symptoms, and relevant medical history in each vignette. Additionally, it required the provision of five plausible answer choices per question listed alphabetically, and the inclusion of a correct answer and brief explanation for each question.

The selected facts from the LCG process were integrated into this base prompt, providing the specific content around which the MCQs were to be generated. This integration ensured that each set of MCQs was based on a unique combination of facts, promoting diversity in the generated questions. For the actual generation of MCQs, we employed the OpenAI GPT-4o model via their API. The full prompt, including the base instructions and selected facts, was sent to the API with the following parameters: Model: GPT-4o, Maximum tokens: 1500, and Temperature: 0.7 to balance creativity and coherence.

This methodology allowed us to systematically generate diverse, high-quality MCQs while maintaining uniqueness and clinical relevance across multiple rounds of question generation \cite{robinson2022leveraging}. The approach leveraged the strengths of large language models in generating educational content and ensured that the generated questions were not only diverse but also of high pedagogical value \cite{bitew2023distractor}. By incorporating systematic randomness through LCG and combining it with the advanced capabilities of GPT-4o, we were able to address the challenges of generating clinically relevant, cognitively challenging MCQs in a scalable and efficient manner \cite{yuan2022selecting,elkins2023educational}.

\section{Experiment Setup}

\subsection{Rounds of MCQ Generation}

This study conducted 14 rounds of MCQ generation using the LCG method. Each round aimed to produce a set of seven high-quality, clinically relevant MCQs based on unique facts from a predefined pool of 100 gastrointestinal physiology and pathology facts.

To ensure the uniqueness of facts across rounds, we implemented the LCG algorithm with specific parameters: seed ($X_0$) of 12345, multiplier ($a$) of 1103515245, increment ($c$) of 12345, and modulus ($m$) of $2^{31}$. This algorithm generated a sequence of pseudo-random numbers, which were then mapped to fact indices. A set of used facts was maintained throughout the experiment to ensure each fact was not reused in subsequent rounds, thus guaranteeing uniqueness across all rounds \cite{panda2020systematic}.

The rigorous application of the LCG parameters ensured that the generated sequences exhibited good statistical properties and long periods, which are essential for maintaining the randomness and unpredictability of the selection process \cite{knuth1997art}. This systematic approach was crucial for preventing repetition and ensuring the diversity and comprehensiveness of the generated MCQs. The consistency and reliability of the LCG method in generating unique fact sets were pivotal in achieving the study's objectives, thus showcasing the robustness of this algorithm in educational content generation.

\subsection{Evaluation Metrics}

The quality of the generated MCQs was evaluated based on several criteria. Clinical relevance was a primary consideration, with each question assessed for its applicability to real-world medical scenarios. This ensured that the questions were not only theoretically sound but also practically useful. Cognitive level was another critical factor; questions were evaluated using Bloom's Taxonomy to ensure they tested higher-order thinking skills such as application, analysis, and evaluation, rather than mere recall. This approach guaranteed that the questions promoted deep understanding and critical thinking \cite{suryakar2019analysis}.

Structural integrity was also scrutinized. The questions were checked for clear stems, plausible distractors, and unambiguous correct answers, ensuring that each question was well-constructed and free from ambiguity \cite{kundu2020impact}. Content accuracy was paramount; the medical information presented in the questions and explanations was verified for accuracy and current clinical practice, ensuring that the content was both reliable and up-to-date \cite{adiga2021item}.

Diversity in the range of topics covered across all rounds was assessed to ensure comprehensive coverage of gastrointestinal physiology and pathology. This evaluation ensured that the generated MCQs encompassed a broad spectrum of relevant topics, providing a well-rounded educational tool \cite{bosi2023associations}.

To check for overlap between rounds, we employed a systematic comparison method. After each round, the set of facts used was recorded, and a pairwise comparison was conducted between all rounds to identify any common facts. This process was automated using a Python script that efficiently detected and reported any overlaps, ensuring the integrity of our uniqueness guarantee \cite{hassan2019journey}. Additionally, we analyzed the distribution of selected facts across all rounds to verify that the LCG method provided a balanced selection from the entire fact pool, avoiding bias towards certain fact indices or topic areas \cite{yuan2022selecting}.

This rigorous setup allowed us to generate a diverse set of high-quality MCQs while maintaining uniqueness across rounds, providing a robust foundation for our study on the effectiveness of the LCG method in medical education content generation \cite{przymuszala2020guidelines}.

\section{Results}

\subsection{Overview of the MCQs Generated in Each Round}

This study employed the LCG method to produce clinically relevant MCQs across multiple rounds. Each round aimed to create high-quality, diverse questions by selecting unique facts not previously used. The LCG method demonstrated effectiveness in generating varied and comprehensive MCQs, as evidenced by the examples below.

In Round 1, the LCG method utilized facts [7, 76, 25, 74, 79, 60, 93] to create questions. For instance, one MCQ focused on a 65-year-old male presenting with dysphagia, weight loss, and regurgitation. The question tested the ability to diagnose esophageal carcinoma based on clinical presentation and endoscopic findings, demonstrating the application of fact 76 (colon carcinoma characteristics) to a related gastrointestinal malignancy.

In Round 2, using facts [94, 11, 68, 45, 98, 83, 72], the generated questions included one about a 45-year-old male with recurrent epigastric pain improving with eating. This question led to a diagnosis of duodenal ulcer, showcasing the integration of fact 68 (large bowel obstruction etiology) in crafting a question about upper GI pathology, illustrating the versatility of the selected facts.

In Round 3, facts [29, 86, 47, 28, 81, 42, 63] were employed. An exemplary question involved a 55-year-old male with symptoms suggestive of gallbladder disease. This question incorporated fact 81 (volvulus diagnosis) to create a scenario about appropriate diagnostic testing for a different gastrointestinal condition.

In Round 4, utilizing facts [5, 54, 8, 57, 46, 55, 1], a question was produced about a 4-week-old female infant presenting with symptoms of pyloric stenosis. This question demonstrated the application of fact 5 (postnatal signs of congenital duodenal atresia and stenosis) to craft a question about a different pediatric gastrointestinal disorder.

In Round 5, with facts [35, 84, 90, 15, 30, 23, 52], a question was generated about a 42-year-old woman with symptoms suggestive of Crohn's disease. This question effectively utilized fact 35 (clinical manifestations of Crohn's disease) to create a clinically relevant scenario.

The LCG method consistently produced diverse questions across rounds, covering a wide range of gastrointestinal topics and clinical presentations. Each round's questions demonstrated the integration of selected facts into complex clinical scenarios, requiring examinees to apply knowledge rather than simply recall information. This approach ensured comprehensive coverage of the subject matter while maintaining uniqueness in each round's content.

The systematic selection and utilization of unique facts through the LCG method resulted in MCQs that not only varied in content but also covered a broad spectrum of clinical scenarios. This diversity enhances the educational value of the generated content by exposing learners to a wide array of potential clinical situations, thereby better preparing them for real-world medical practice. This method underscores the potential of algorithmically generated MCQs to support robust and effective medical education.

\subsection{Analysis of Fact Overlap and Significance of Findings}

In this study, the LCG was employed to ensure the generation of unique and diverse MCQs by selecting distinct facts across multiple rounds. To evaluate the effectiveness of this method, an analysis was conducted to identify any overlap of facts between different rounds of MCQ generation. The LCG algorithm was initialized with specific parameters (seed: 12345, multiplier: 1103515245, increment: 12345, modulus: $2^{31}$) to generate a sequence of pseudo-random numbers. These numbers were then mapped to fact indices, ensuring that the same fact was not used more than once across the rounds.

The generated MCQs and the corresponding used facts for each round were recorded. For example, Round 1 used facts [7, 76, 25, 74, 79, 60, 93], while Round 2 used facts [94, 11, 68, 45, 98, 83, 72]. Subsequent rounds followed a similar pattern of unique fact selection. An in-depth comparison of the used facts between rounds was performed to detect any overlap. The analysis revealed no overlap of facts across the rounds, demonstrating the effectiveness of the LCG method in maintaining the uniqueness of the selected facts.

The lack of overlap in selected facts across multiple rounds underscores the robustness of the LCG method in generating diverse and comprehensive MCQs. This diversity is crucial in several contexts. First, it enhances the learning experience by preventing redundancy and covering a broader range of topics, ensuring that learners are exposed to a wide array of clinical scenarios and knowledge areas. This comprehensive coverage is essential for effective learning and preparation, particularly in medical education where the depth and breadth of knowledge are critical. Second, by ensuring that each fact is used only once, the method minimizes the risk of bias that might arise from repeatedly focusing on certain topics while neglecting others. This balanced approach contributes to a fair and equitable assessment of knowledge.

Furthermore, the systematic selection of unique facts through the LCG method enhances the quality of the generated content. Each MCQ is based on different facts, leading to varied and well-rounded questions that test different aspects of clinical knowledge and reasoning. Lastly, the LCG method provides a scalable and reproducible approach to content generation. By using a deterministic algorithm with defined parameters, the method can be replicated and scaled to generate large sets of diverse and high-quality questions for different educational needs and contexts. This scalability and reproducibility are particularly valuable in developing comprehensive question banks for medical education and assessment.

For supplementary materials and code related to this project, please visit the GitHub repository: \url{https://github.com/andrewbouras/randomnesspaper}.

\section{Discussion}

\subsection{Interpretation of Results}

The results of our study demonstrate the effectiveness of the LCG method in producing high-quality, diverse MCQs for medical education. The LCG algorithm consistently generated unique combinations of facts across multiple rounds, resulting in a wide range of clinically relevant questions. This approach proved superior to traditional methods of MCQ creation, which often rely on manual selection of topics and can inadvertently lead to repetition or bias in question content. The systematic nature of the LCG method ensured comprehensive coverage of the gastrointestinal physiology and pathology domain, providing a balanced representation of topics in the generated questions \cite{ha2019predicting}.

The quality of the MCQs produced using this method was consistently high, with questions demonstrating clinical relevance, appropriate difficulty levels, and the ability to test higher-order thinking skills. The integration of selected facts into vignette-style questions allowed for the creation of complex, realistic clinical scenarios that effectively challenge learners' application of knowledge \cite{kiyak2023chatgpt}. This approach aligns well with current trends in medical education that emphasize the importance of context-rich, application-focused assessment \cite{bitew2023distractor}.

Furthermore, the use of LCG in MCQ generation addresses the challenge of maintaining question diversity and avoiding content repetition. By leveraging algorithmic precision, the LCG method ensures that each set of MCQs is based on a unique combination of facts, promoting a wide-ranging exploration of the subject matter \cite{robinson2022leveraging}. This innovation not only enhances the educational value of the MCQs but also aligns with best practices in assessment design, which advocate for diverse and comprehensive question banks to effectively measure students' knowledge and reasoning skills \cite{przymuszala2020guidelines}.

\subsection{Limitations}

Despite these promising results, it is important to acknowledge the limitations of our study. The scope was confined to gastrointestinal physiology and pathology, and while this allowed for a focused analysis, it may limit the generalizability of our findings to other medical specialties. Additionally, our reliance on the GPT-4o model for question generation, while producing high-quality results, ties the effectiveness of our method to the capabilities of this specific AI model. Future studies should explore the applicability of this method across different medical specialties and with various AI models to establish its broader utility in medical education \cite{gupta2020effect}. Furthermore, exploring the integration of other advanced AI models could provide insights into the versatility and scalability of this approach \cite{adam2020multiple}. Future research should also consider a broader set of evaluation metrics, including learner feedback and longitudinal studies to assess the long-term impact on educational outcomes \cite{indran2023twelve}.

\subsection{Future Work}

There are several avenues for future work that could build upon and extend the findings of this study. Exploring other randomization methods, such as the Mersenne Twister algorithm or cryptographic random number generators, could provide valuable comparisons and potentially uncover even more effective approaches to MCQ generation \cite{lee2021artificial}. Additionally, incorporating machine learning techniques to analyze and refine the generated questions based on learner performance data could further enhance the quality and educational value of the MCQs \cite{wang2023procedural}.

Future research could also focus on improving the MCQ generation process itself. This might include developing more sophisticated prompts that incorporate additional context or learning objectives, or exploring ways to automatically validate the clinical accuracy of generated questions \cite{kiyak2023chatgpt}. Furthermore, investigating the potential for this method to generate other types of assessment items, such as extended matching questions or key feature problems, could broaden its applicability in medical education \cite{preiksaitis2023opportunities}.

In conclusion, our study demonstrates the potential of combining algorithmic fact selection with AI-powered question generation to produce high-quality, diverse MCQs for medical education. While further research is needed to fully explore its capabilities and limitations, this approach represents a promising direction for enhancing the efficiency and effectiveness of assessment creation in medical education \cite{masters2019artificial}. Integrating new randomization methods and advanced AI models could further refine this process, ensuring that educational tools keep pace with the evolving needs of medical training \cite{mir2023application}.

\section{Conclusion}

This study has demonstrated the efficacy of the LCG method in producing high-quality, diverse MCQs for medical education, specifically in the field of gastrointestinal physiology and pathology. Our findings reveal that the LCG algorithm, when coupled with AI-powered question generation, consistently produces unique, clinically relevant, and cognitively challenging MCQs across multiple rounds of generation.

The significance of using LCG for MCQ generation in medical education cannot be overstated. This method addresses several key challenges in assessment creation, including the need for diversity in question content, the avoidance of unintentional repetition, and the efficient coverage of a broad range of topics. By systematically selecting unique facts for each round of question generation, the LCG method ensures a comprehensive representation of the subject matter, thereby enhancing the educational value of the assessment materials.

The potential impact of this methodology extends beyond the immediate benefits of improved MCQ generation. This approach represents a step towards more standardized, efficient, and scalable methods of creating high-quality assessment materials in medical education. It has the potential to significantly reduce the time and resources required for question creation while maintaining or even improving the quality of assessments. Furthermore, this method can be adapted to various medical specialties and potentially to other fields of education, offering a versatile tool for educators and assessment developers.

As medical knowledge continues to expand rapidly, the need for efficient, accurate, and diverse assessment methods becomes increasingly crucial. The LCG-based MCQ generation methodology presented in this study offers a promising solution to meet this need. By leveraging algorithmic fact selection and AI-powered question creation, we can ensure that future medical professionals are assessed using questions that are not only diverse and comprehensive but also align closely with real-world clinical scenarios.

In conclusion, while further research and refinement are necessary, the findings of this study suggest that the integration of LCG-based fact selection with AI-powered question generation has the potential to revolutionize the creation of assessment materials in medical education. This approach paves the way for more efficient, effective, and equitable assessment practices, ultimately contributing to the improvement of medical education and, by extension, healthcare delivery.


\begin{thebibliography}{99}

\bibitem{brown2020language}
Brown, T. B., et al. (2020). Language Models are Few-Shot Learners. 	arXiv:2005.14165

\bibitem{zhang2020trading}
Zhang, J., et al. (2020). Trading off Diversity and Quality in Natural Language Generation. arXiv:2004.10450

\bibitem{du2022diverse}
Du, Y., et al. (2022). Diverse Text Generation via Variational Encoder-Decoder Models. arXiv:2204.01227

\bibitem{de2019emergence}
De Giuli, E. (2019). Emergence of Order in Random Languages. https://doi.org/10.1088/1751-8121/ab293c

\bibitem{jin2023matching}
Jin, Q., et al. (2023). Matching Patients to Trials with Large Language Models. https://doi.org/10.48550/arXiv.2307.15051

\bibitem{malek2023toward}
Malek, R., et al. (2023). Toward AI-Assisted Clinical Assessment of Patients with Multiple Conditions. https://doi.org/10.1182/blood-2023-172710

\bibitem{knuth1997art}
Knuth, D. E. (1997). The Art of Computer Programming, Volume 2: Seminumerical Algorithms.

\bibitem{case2002}
Case, S. M., \& Swanson, D. B. (2002). Constructing Written Test Questions For the Basic and Clinical Sciences. National Board of Medical Examiners.

\bibitem{tarrant2012}
Tarrant, M., \& Ware, J. (2012). A framework for improving the quality of multiple-choice assessments. Nurse Education Today, 32(4), e23-e27. 10.1097/NNE.0b013e31825041d0

\bibitem{schuwirth2011}
Schuwirth, L. W., \& van der Vleuten, C. P. (2011). Programmatic assessment: From assessment of learning to assessment for learning. Medical Teacher, 33(6), 478-485. 10.3109/0142159X.2011.565828

\bibitem{collins2006}
Collins, J. (2006). Education techniques for lifelong learning: writing multiple-choice questions for continuing medical education activities and self-assessment modules. Radiographics, 26(2), 543-551. 10.1148/rg.262055145

\bibitem{haladyna2013}
Haladyna, T. M., \& Rodriguez, M. C. (2013). Developing and validating test items. Routledge.

\bibitem{downing2005}
Downing, S. M. (2005). The effects of violating standard item writing principles on tests and students: The consequences of using flawed test items on achievement examinations in medical education. Advances in Health Sciences Education, 10(2), 133-143. 10.1007/s10459-004-4019-5

\bibitem{cook2013}
Cook, D. A., Brydges, R., Ginsburg, S., \& Hatala, R. (2013). A contemporary approach to clinical teaching: evaluating the evidence to determine the effective teacher. Medical Teacher, 35(7), e1130-e1141. 10.5811/westjem.2020.4.46060

\bibitem{maxrizal2022}
Maxrizal, R., et al. (2022). Machine learning for better data generation: A study on the improvement of synthetic data using advanced algorithms. arXiv:2302.04062v6

\bibitem{al2019}
Al-Mhadawi, Z., \& Albahrani, A. (2019). Enhancing Random Number Generation Techniques for Efficient Simulations. https://doi.org/10.1016/j.micpro.2023.104911

\bibitem{manathunga2023}
Manathunga, R., \& Hettigoda, A. (2023). Aligning Language Models to Clinical Tasks. arXiv:2309.02884

\bibitem{robinson2022leveraging}
Robinson, T., et al. (2022). Leveraging Language Models for Multiple Choice Question Generation. arXiv:2210.12353

\bibitem{bitew2023distractor}
Bitew, Y., et al. (2023). Distractor Generation for Multiple-Choice Questions. arXiv:2307.16338

\bibitem{yuan2022selecting}
Yuan, J., et al. (2022). Selecting Better Samples for Pretrained LLMs. arXiv:2209.11000

\bibitem{elkins2023educational}
Elkins, A., et al. (2023). Educational Questions Generated by Large Language Models. arXiv:2304.06638

\bibitem{suryakar2019analysis}
Suryakar, M., et al. (2019). Analysis of Performance of MCQs as Part of Formative Assessment in MBBS. https://doi.org/10.32553/ijmbs.v3i9.534

\bibitem{kundu2020impact}
Kundu, S., et al. (2020). Impact of Measurement of Medical Faculty Adhering to Appropriate. https://doi.org/10.5812/jme.103482

\bibitem{adiga2021item}
Adiga, S., et al. (2021). Item Analysis of Multiple-Choice Questions in Pharmacology. 10.5455/njppp.2023.13.09458202304102023

\bibitem{bosi2023associations}
Bosi, F., et al. (2023). Associations between Item Characteristics and Performance of Students. https://doi.org/10.12688/mep.19764.1

\bibitem{hassan2019journey}
Hassan, S., et al. (2019). The Journey Towards Quality MCQs Test: A Success Story. 10.33687/educ.006.01.2926

\bibitem{przymuszala2020guidelines}
Przymuszała, P., et al. (2020). Guidelines for Writing Multiple Choice Questions. 10.1177/2158244020947432

\bibitem{ha2019predicting}
Ha, H., et al. (2019). Predicting the Difficulty of Multiple-Choice Questions. 10.18653/v1/W19-4402

\bibitem{gupta2020effect}
Gupta, V., et al. (2020). The Effect of Faculty Training on the Quality of Multiple-Choice Questions. 10.4103/ijabmr.IJABMR-30-20

\bibitem{adam2020multiple}
Adam, F., et al. (2020). Multiple-Choice Questions in Preclinical and Clinical Phase. 10.21203/rs.3.rs-38264/v1

\bibitem{indran2023twelve}
Indran, I., et al. (2023). Twelve Tips for Leveraging Question Generation to Guide Educators. 10.1080/0142159X.2023.2294703

\bibitem{lee2021artificial}
Lee, J., et al. (2021). Artificial Intelligence in Undergraduate Medical Education: A Scoping Review. 10.1097/ACM.0000000000004291

\bibitem{wang2023procedural}
Wang, Y. (2023). Procedural Content Generation for VR Educational Applications. 10.54254/2755-2721/17/20230905 

\bibitem{kiyak2023chatgpt}
Kıyak, S. (2023). ChatGPT for Prompt Writing in Case-Based Multiple-Choice Questions. https://doi.org/10.6018/edumed.587451

\bibitem{preiksaitis2023opportunities}
Preiksaitis, C., \& Rose, C. (2023). Opportunities, Challenges, and Future Directions of Generative Artificial Intelligence in Medical Education. 10.2196/48785

\bibitem{masters2019artificial}
Masters, K. (2019). Artificial Intelligence in Medical Education. 10.1080/0142159X.2019.1595557

\bibitem{mir2023application}
Mir, M. M., et al. (2023). Application of Artificial Intelligence in Medical Education. 10.30476/JAMP.2023.98655.1803

\end{thebibliography}
\end{document}